\documentclass[10pt,twocolumn,letterpaper]{article}

\usepackage[pagenumbers]{cvpr} 
\usepackage[accsupp]{axessibility}
\usepackage[dvipsnames]{xcolor}

\usepackage{multirow}
\definecolor{cvprblue}{rgb}{0.21,0.49,0.74}
\definecolor{lightblue}{RGB}{149,174,222}
\definecolor{lightgreen}{RGB}{174,212,149}
\usepackage[pagebackref,breaklinks,colorlinks,citecolor=cvprblue]{hyperref}
\def\paperID{5445} 
\def\confName{CVPR}
\def\confYear{2024}
\usepackage[misc]{ifsym}
\usepackage[hang]{footmisc}

\newcommand\blfootnote[1]{
    \begingroup
    \renewcommand\thefootnote{}\footnote{#1}
    \addtocounter{footnote}{-1}
    \endgroup
}

\title{Sparse Global Matching for Video Frame Interpolation with Large Motion}

\author{Chunxu Liu$^{1, *}$ \quad Guozhen Zhang$^{1, *}$ \quad Rui Zhao$^2$ \quad Limin Wang$^{1, 3, \textrm{\Letter}}$\\
$^1$State Key Laboratory for Novel Software Technology, Nanjing University \\ $^2$ SenseTime Research \quad \quad $^3$ Shanghai AI Lab \\
\textbf{\normalsize\url{https://sgm-vfi.github.io}}
}

\begin{document}
\maketitle
\begin{abstract}
Large motion poses a critical challenge in Video Frame Interpolation (VFI) task. Existing methods are often constrained by limited receptive fields, resulting in sub-optimal performance when handling scenarios with large motion. In this paper, we introduce a new pipeline for VFI, which can effectively integrate global-level information to alleviate issues associated with large motion. Specifically, we first estimate a pair of initial intermediate flows using a high-resolution feature map for extracting local details. Then, we incorporate a sparse global matching branch to compensate for flow estimation, which consists of identifying flaws in initial flows and generating sparse flow compensation with a global receptive field. Finally, we adaptively merge the initial flow estimation with global flow compensation, yielding a more accurate intermediate flow. To evaluate the effectiveness of our method in handling large motion, we carefully curate a more challenging subset from commonly used benchmarks. Our method demonstrates the state-of-the-art performance on these VFI subsets with large motion.
\end{abstract}

\blfootnote{*: Equal Contribution.~~~\Letter: Corresponding author (lmwang@nju.edu.cn).}
\vspace{-5mm}
\section{Introduction}
\begin{figure}[t]
  \centering
  \includegraphics[width=1.0\linewidth, trim=0cm 6cm 0cm 1cm, clip=true]{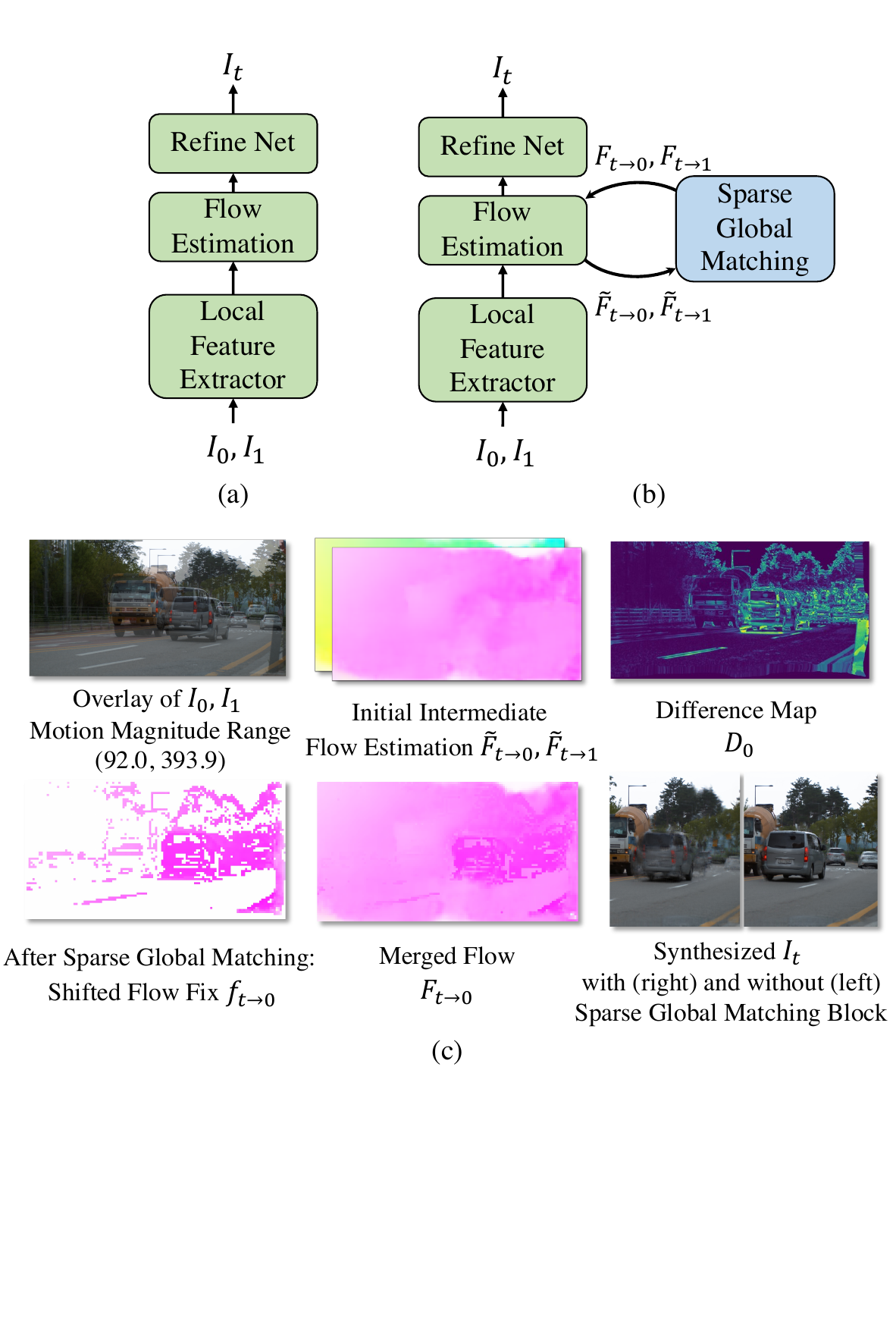}
  \vspace{-5mm}
  \caption{(a) \textbf{Our framework without sparse global matching}, pretrained on small motion dataset, for capturing local details. (b) \textbf{Our framework with sparse global matching}, fine-tuned on large motion dataset, for capturing global large motion. (c) \textbf{Key components in our algorithm}, illustrating the effect of our sparse global matching branch. (Using Ours-1/4-Points, from \Cref{tab:main}.)}
  \label{fig:introfig}
  \vspace{-4mm}
\end{figure}

Video Frame Interpolation (VFI) seeks to generate the intermediate frame from a given pair of inference frames, which has received increasing attention. It has various real-life applications, such as creating slow motion videos~\cite{slomo, safa, videoinr}, video compression~\cite{NCM, vidcom} and novel view synthesis~\cite{LearningTo, deepstereo, blur}. 
Currently, flow-based algorithms occupy a prominent position~\cite{EMA, RIFE, BiFormer, VFIT, IFRNet, AMT, dav} in VFI task, where the flow from the target frame to the input frames, namely, intermediate flow, is explicitly estimated for warping input frames to the target frame. 

Nevertheless, existing optical flow~\cite{RAFT, gmflow, liteflownet} algorithms cannot be directly applied to estimate intermediate flows due to the absence of the target frame.
To address this, many algorithms first estimate the bidirectional flow between two input frames and then generate intermediate optical flows using various flow reversal techniques~\cite{QVI, EQVI, slomo, XVFI, BMBC, zhou2023, upr}. 
Alternatively, recent algorithms directly estimate the intermediate flow with proper supervision~\cite{IFRNet, EMA, RIFE} and achieve improved performance on datasets where small-to-medium motions are prevalent~\cite{FILM}. 

However, real-world video frame interpolation encounters various complex challenges, with the problem of handling large motion being particularly prominent. 
In scenarios characterized by large motion, the correspondence of objects between frames is hard to locate due to the large pixel shifting. 
Many works have been proposed to alleviate this problem. For example, XVFI~\cite{XVFI} addressed this by creating an extreme 4K training dataset and a scalable framework. However, its performance improvement is limited when applied to small motion scenarios~\cite{bilat}. FILM~\cite{FILM} introduced scale-agnostic features and a weight-sharing approach to improve model generalization across different motion scales. In fact, due to its limited receptive field, the model's depth can become excessively deep when dealing with large motion. This increases computational complexity and falls short in handling fast-moving small objects.

In this paper, we introduce a sparse global matching pipeline to specifically handle the challenges posed by large motion in the VFI task, by effectively integrating global-level information into intermediate flow estimation. 
Our VFI method establishes sparse global correspondences between input frames using a global receptive field and takes a two-step strategy to compensate for the intermediate flow estimation. 
Specifically, as shown in Figure~\ref{fig:introfig} (a - b), our method begins with an initial estimation of a pair of intermediate flows using a relatively high-resolution feature map. Following this initial estimation, our approach incorporates a sparse global matching branch to locate potential error in the flow estimation results, and then produce flow residual to provide an effective remedy for capturing large motion. 

To be specific, in our sparse global matching branch, we build a difference map to pinpoint the flaws in initial intermediate flow estimations. Concentrating on these defective areas, our approach employs sparse global matching to establish global correspondences between two adjacent input frames, specifically at these sparsely targeted locations. Subsequently, we convert this bidirectional sparse flow correspondence into intermediate flow compensation. Finally, we employ a flow merging block to adaptively merge the initial intermediate flows and flow compensation, thereby effectively combining the local details with the global correlation. As shown in~\Cref{fig:introfig} (c), our sparse global matching branch can effectively locate the error regions and rectify the flaws in the initial intermediate flow estimation, yielding a significantly enhanced synthesized intermediate frame.

In order to benchmark the effectiveness of our sparse global matching module on handling large motion, we carefully analyze motion magnitude and sufficiency within existing benchmarks, X-Test~\cite{XVFI}, Xiph~\cite{xiph} and SNU-FILM~\cite{CAIN}. In our analysis, we utilize the motion sufficiency filtering method described in~\cite{LearningTo}, emphasizing the minimum of the top 5\% of each pixel's flow magnitude as the key indicator of large motion sufficiency. In the end, we carefully curate the most challenging subsets for large motion frame interpolation evaluation. In the most challenging testing conditions, our proposed method demonstrates a substantial improvement in terms of PSNR, enhancing it by \textbf{$0.66$} dB while correcting half of the points in initial flow estimation. Furthermore, even with the correction of just $1/8$ points, we still observe a notable increase of \textbf{$0.48$} dB. Notably, our approach establishes a new state-of-the-art performance benchmark in these exceptionally challenging scenarios. In summary, our main contributions include: 
\begin{itemize}
    \item We introduce a sparse global matching algorithm tailored to effectively capture large motion.
    \item We designed an effective two-step framework for capturing large motion. First, by estimating initial intermediate flows to extract the local details, then targets and corrects the detected flaws through sparse global matching at sparsely targeted defective points.
    \item Our models demonstrate state-of-the-art performance on the most challenging subset of the commonly used large motion benchmark, namely, X-Test-L, Xiph-L, SNU-FILM-L hard and extreme.
\end{itemize}
\section{Related Work}
\label{sec:rw}

\begin{figure*}[t]
    \centering
    \includegraphics[width=\linewidth, trim=6cm 5.5cm 4.3cm 6.5cm, clip=true]{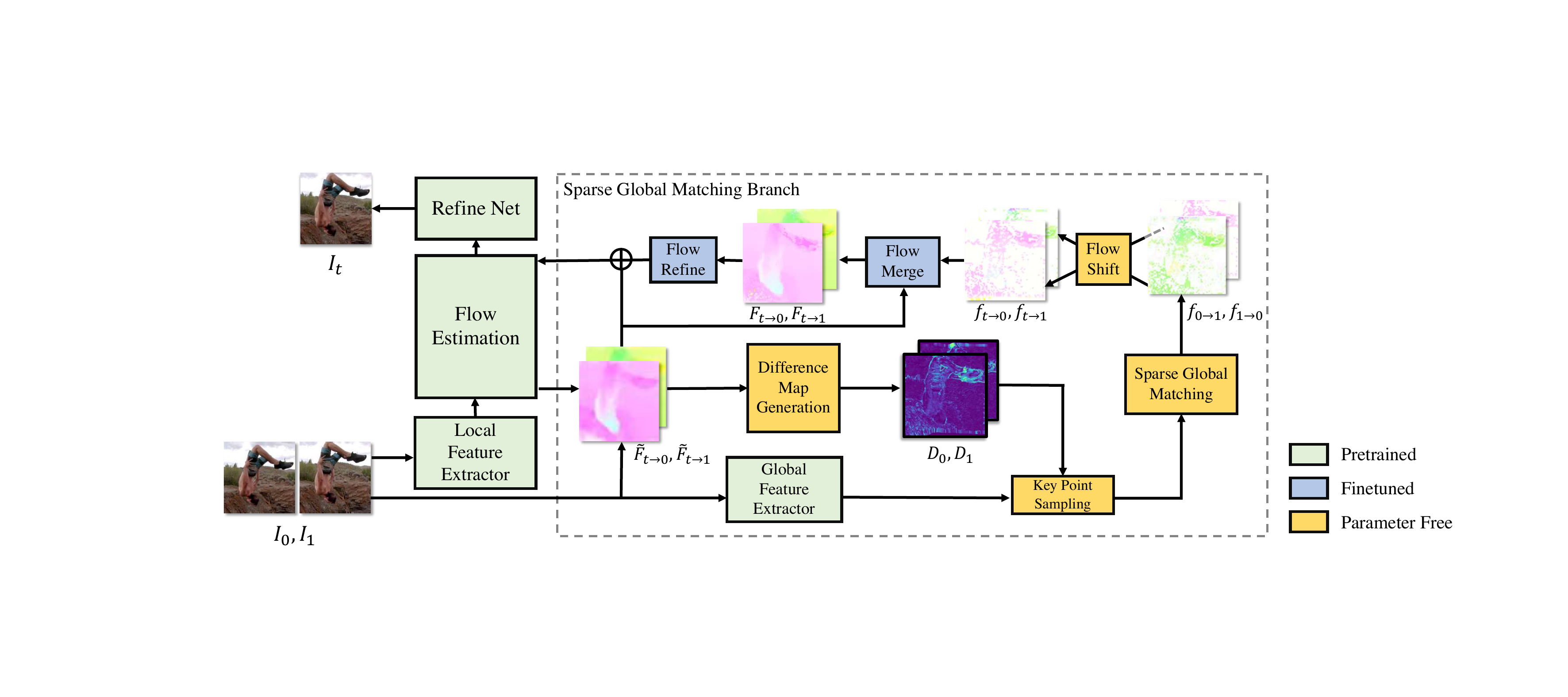}
    \caption{\textbf{Overview of our proposed structure.} First, local features are extracted by a local feature extractor for flow estimation $\widetilde{F}_{t\rightarrow 0}, \widetilde{F}_{t\rightarrow 1}$. Then, our sparse global matching branch locates the flaws by constructing difference maps $D_0, D_1$. Next, we perform sparse global matching using global features extracted by a global feature extractor. Finally, after shifting global correspondences $f_{0\rightarrow 1}, f_{1\rightarrow 0}$ to intermediate flow compensation $f_{t\rightarrow 1}, f_{t\rightarrow 0}$, we adaptively merge $\widetilde{F}_{t\rightarrow 0}, \widetilde{F}_{t\rightarrow 1}$ with $f_{t\rightarrow 1}, f_{t\rightarrow 0}$ and adopts a flow refine in a residual manner for interpolating the intermediate frame.}
    \label{fig:pipeline}
    \vspace{-4mm}
\end{figure*}

\subsection{Flow-Based Video Frame Interpolation}
Flow-based algorithms for video frame interpolation focus on estimating intermediate flows. These flows enable the model to either forward warp or backward warp the input frames to the target frame. Some algorithms first compute bidirectional flows between two input frames and apply different flow reversal techniques to obtain intermediate flows~\cite{BMBC, QVI, XVFI}, while the others directly estimate intermediate flows~\cite{RIFE, EMA, IFRNet}.

However, flow reversal techniques may introduce artifacts to intermediate flows, which can harm the details of the intermediate flows, causing misalignment~\cite{slomo} or holes~\cite{QVI}. Direct estimation of intermediate flows is also restricted by the model design, usually has a limited perceptive field~\cite{RIFE, EMA}, and lacks robustness on large motion.

Addressing the issue of large motion in video frame interpolation, FILM~\cite{FILM} adopts a coarse-to-fine approach and shares weights between layers. It also trains on data with varied motion magnitudes, enabling it to learn to handle large motions. AMT~\cite{AMT} uses the RAFT-like structure to construct an all-pair correlation map and locally refines the intermediate flow afterward. BiFormer~\cite{BiFormer} employs a global feature extractor to extract global features and build local bilateral correlation maps. However, due to the high resolution of the VFI task, giving the model a genuine global receptive field is challenging.

In contrast, our work uses local features to estimate the intermediate flows and global features to sparsely generate bidirectional flow compensation by global matching. We then adaptively merge the flows and flow compensation to obtain the final intermediate flow with both global information and local details.

\subsection{Transformer}
Transformers have gained widespread popularity across various computer vision tasks~\cite{vit, swin, detr,ddetr}, demonstrating impressive feature extraction capabilities. Recently, Transformer-based backbone has already been introduced to solve VFI task~\cite{xiph,cavfi, VFIT, EMA}. VFIFormer~\cite{VFIFormer} uses cross-scale window-based attention. EMA-VFI~\cite{EMA} uses Inter-frame Attention to extract appearance and motion features using the same similarity matrix. BiFormer~\cite{BiFormer} adopts pretrained Twins architecture~\cite{twins} for global feature extraction and builds local bilateral correlation cost volume. However, when handling large motion, these methods are more or less restricted by their local receptive field. Therefore, we propose a two-step strategy: employing a hybrid Transformer and CNN-based backbone for initial flow estimation, followed by a sparse global matching block utilizing a global receptive field.

\subsection{Global Matching}
Global matching is extensively studied in local feature matching tasks. LoFTR~\cite{LoFTR} replaces the traditional dense matching method, using cost volume to search for correspondence, with self and cross attention layers in Transformer. COTR~\cite{COTR} takes a different approach by formulating the correspondence problem as a functional mapping and utilizing a Transformer as the function to query the points of interest. For dense matching, GMFlow~\cite{gmflow} adopted Transformer block to extract strong discriminative features to build a global matching map. In VFI tasks, not every part of the image requires dense global matching. Our sparse global matching strategy, on the other hand, specifically targets and corrects the defective areas of the flows.

\section{Method}
Given a pair of RGB frames $I_0$, $I_1$, and a timestep $t$, we need to synthesize an intermediate frame $I_t$. We illustrate our overall model pipeline in \Cref{fig:pipeline}. 

Our method consists of a local feature branch and a sparse global matching branch. The local feature branch is responsible for estimating the initial intermediate flows, namely $\widetilde{F}_{t\rightarrow 0}$ and $\widetilde{F}_{t\rightarrow 1}$. In the sparse global matching branch, we focus on the defective areas of $\widetilde{F}_{t\rightarrow 0}$, $\widetilde{F}_{t\rightarrow 1}$, and perform sparse global matching to obtain flow compensation, $f_{t\rightarrow 0}$ and $f_{t\rightarrow 1}$. Then we adaptively merge $\widetilde{F}_{t\rightarrow 0}$ and $\widetilde{F}_{t\rightarrow 1}$ with $f_{t\rightarrow 0}$ and $f_{t\rightarrow 1}$ by our flow merge block, to obtain more accurate intermediate flows, $F_{t\rightarrow 0}$, $F_{t\rightarrow 1}$. Finally, after a few flow refine blocks, we use this pair of intermediate flows to synthesize the intermediate frame $\widehat{I_t}$. 

In the following, we first introduce each component in our local feature branch briefly in \Cref{sec:main-branch}. From \Cref{sec:diff-map} and onward, we delve into a detailed discussion of our sparse global matching branch.
\subsection{Local Feature Branch}
\label{sec:main-branch}
\label{sec:method}

As shown in \Cref{fig:pipeline}, our local feature branch consists of three parts, namely, the local feature extractor, the flow estimation block, and the RefineNet~\cite{RIFE}. We draw inspiration from RIFE~\cite{RIFE} and EMA-VFI~\cite{EMA}, while retaining certain distinctive aspects. We now introduce them sequentially.

\noindent \textbf{Local Feature Extractor. }EMA-VFI~\cite{EMA} has already verified the effectness of CNN and Transformer hybrid structure. We follow its design and keep our deepest feature resolution stays at $H/8 \times W/8$ scale, instead of $H/16 \times W/16$ in EMA-VFI. Furthermore, we simplified the specially designed Transformer structure in EMA-VFI to simple cross attention to extract inter-frame appearance features at a lower computation cost. 

\noindent \textbf{Flow Estimation.} We use the extracted appearance feature and input frames $I_0, I_1 \in \mathbb{R}^{3\times H\times W}$ to directly estimate intermediate flows $F_{t\rightarrow 0}$, $F_{t\rightarrow 1}$, along with a fusion map $O$, in a coarse-to-fine manner. This is achieved by several layers of CNN block, inspired by the intuitive design of RIFE~\cite{RIFE} and EMA-VFI~\cite{EMA}. Then, the target frame can be generated by:
\begin{align}
    \widehat{I}_t=O \odot \overleftarrow{\mathcal{W}}\left(I_0, F_{t \rightarrow 0}\right)+(1-O) \odot \overleftarrow{\mathcal{W}}\left(I_1, F_{t \rightarrow 1}\right), \label{eq:image-fusion}
\end{align}
where $\overleftarrow{\mathcal{W}}$ is image backward warping operation, $\odot$ is Hadamard product operator. 

\noindent \textbf{Flow Refine.} We share the same U-net-shaped network with RIFE to refine the synthesized frame $\widehat{I}_t$. More details can be found in \Cref{local structure}.

\subsection{Locate Flaws in Flows: Difference Map}
\label{sec:diff-map}
Due to the limited receptive field of local feature branch, the estimated initial bidirectional flows, $\widetilde{F}_{t\rightarrow 0}$ and $\widetilde{F}_{t\rightarrow 1}$, may be relatively coarse. Consequently, it is necessary to identify the flaws in $\widetilde{F}_{t\rightarrow 0}$ and $\widetilde{F}_{t\rightarrow 1}$. Therefore, we construct the difference map $D_0$ and $D_1$ to check the correctness of $\widetilde{F}_{t\rightarrow 0}$ and $\widetilde{F}_{t\rightarrow 1}$, by using both of the input frames $I_0$ and $I_1$ as ground truth reference. The values in the difference maps $D_0$ and $D_1$ serve as indicators of the likelihood of error in flow estimation, with higher values suggesting a greater probability of inaccuracies.

In particular, we first use $\widetilde{F}_{t\rightarrow 0}$ to backward warp $I_0$ to $\widetilde{I}_t$, then use $\widetilde{F}_{t\rightarrow 1}$ to forward warp $\widetilde{I}_t$ to $\widetilde{I}_1$. And we compare the $\widetilde{I}_1$ and input frame $I_1$ by doing summation over RGB channels after subtraction to obtain difference map $\widetilde{D}_{0\rightarrow 1}$. 
\vspace{-2mm}
\begin{align}
    \widetilde{I}_1 &= \overrightarrow{\mathcal{W}} \left( \overleftarrow{\mathcal{W}}\left(I_0, \widetilde{F}_{t\rightarrow 0}\right), \widetilde{F}_{t\rightarrow 1}\right), \\
    \widetilde{D}_{0\rightarrow 1} &= \sum_{RGB}|I_1 - \widetilde{I}_1|,
\end{align}
where $\overleftarrow{\mathcal{W}}$ is backward warp, $\overrightarrow{\mathcal{W}}$ is forward warp.

Through the combination of backward warp and forward warp, we get an initial difference map $\widetilde{D}_{0\rightarrow 1}$. Currently, the flaws in $\widetilde{D}_{0\rightarrow 1}$ are caused by two reasons, the first reason is due to the flaws in our coarse flow estimation, $\widetilde{F}_{t\rightarrow 0}$, $\widetilde{F}_{t\rightarrow 1}$. The second reason is that even if $\widetilde{F}_{t\rightarrow 0}$, $\widetilde{F}_{t\rightarrow 1}$ were perfectly accurate, occlusions and cropping would still occur inevitably, i.e. some existing pixels in $I_0$ disappearing in $I_1$, creating incorrect pixel mappings and holes in the warped image.

To filter out the second cause of flaws, we create a map full of ones, and repeat the above warping process to obtain a $0/1$ mask, $M_{0\rightarrow 1}^{holes}$. The positions in mask $M_{0\rightarrow 1}^{holes}$ where the element becomes 0 correspond to the potential hole areas in the warped image $\widetilde{I}_1$. 

Then, we can multiply this map $M_{0\rightarrow 1}^{holes}$ with our initial difference map $\widetilde{D}_{0\rightarrow 1}$ to filter out the potential holes caused by occlusion and cropping, and obtain $D_{0\rightarrow 1}$, focusing on the flaws caused by inaccurate intermediate flow estimation.
\vspace{-5mm}
\begin{align}
    D_{0\rightarrow 1} = M_{0\rightarrow 1}^{holes} \odot \widetilde{D}_{0\rightarrow 1}.
\end{align}

As a result, $D_{0\rightarrow 1}$ enables us to identify the misaligned regions in $\widetilde{I}_1$ that are caused solely by the flaws in the estimated flows $\widetilde{F}_{t\rightarrow 0}$, $\widetilde{F}_{t\rightarrow 1}$. Furthermore, we  reverse the previous warping combination to find the underlying source points responsible for these misalignments in $\widetilde{I}_1$. 
\vspace{-1mm}
\begin{align}
    D_0  = \overrightarrow{\mathcal{W}} \left( \overleftarrow{\mathcal{W}}\left(D_{0\rightarrow 1}, \widetilde{F}_{t\rightarrow 1}\right), \widetilde{F}_{t\rightarrow 0}\right). 
\end{align}

Difference map $D_0$ illustrates the extent to which each point in $I_0$ leads to the misalignment in $\widetilde{I}_1$. To address this, it is essential to prioritize the regions with higher values in $D_0$. These regions indicate the points that exacerbate significant misalignment. Difference map $D_1$ can be produced symmetrically.

\begin{figure}[t]
    \centering
    \begin{subfigure}[b]{0.49\linewidth}
        \centering
        \includegraphics[width=\linewidth, trim=2cm 3cm 1cm 3cm, clip=true]{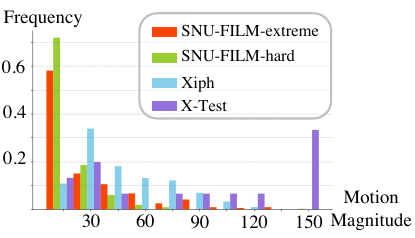}
        \label{fig:c1}
    \end{subfigure}
    \hfill
    \begin{subfigure}[b]{0.49\linewidth}
        \centering
        \includegraphics[width=\linewidth, trim=1.8cm 3cm 1cm 3cm, clip=true]{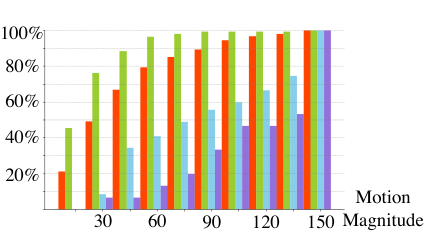}
        \label{fig:c2}
    \end{subfigure}
    
    \vspace{-0.5cm}
    \begin{subfigure}[b]{0.49\linewidth}
        \centering
        \includegraphics[width=\linewidth, trim=2cm 3cm 1cm 3cm, clip=true]{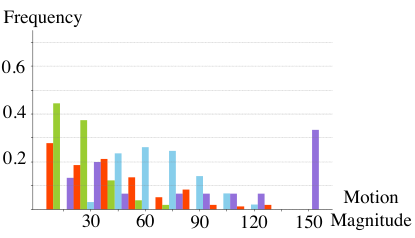}
        \caption{Mean Motion Magnitude}
        \label{fig:c3}
    \end{subfigure}
    \hfill
    \begin{subfigure}[b]{0.49\linewidth}
        \centering
        \includegraphics[width=\linewidth, trim=1.8cm 3cm 1cm 3cm, clip=true]{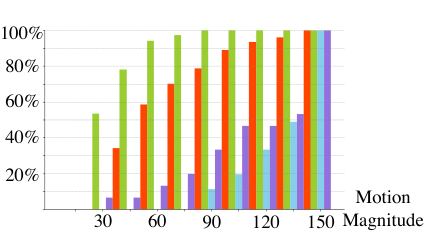}
        \caption{Motion Sufficiency Ranking}
        \label{fig:c4}
    \end{subfigure}

    \caption{\textbf{Large motion dataset benchmark analysis.} \textit{Top:} Whole dataset. \textit{Below:} Keeping the most challenging half of Xiph and SNU-FILM. Four charts share the same legend.}
    \label{fig:charts}
\vspace{-4mm}
\end{figure}

\subsection{Sparse Global Matching}
To emphasize the region with higher values in $D_0$ and $D_1$, we can select the top-$k$ points out of $D_0$, $D_1$ and do global matching for those points to provide more accurate matching flow $f_{0\rightarrow 1}$ and $f_{1\rightarrow 0}$, for preparing the flow compensation $f_{t\rightarrow 1}$ and $f_{t\rightarrow 1}$. 

Sparse global matching allows the selected points in $I_0$ to have a global receptive field on $I_1$. We adopt the pretrained global feature extractor from GMFlow~\cite{gmflow} to generate discriminative global features $A_0^i, A_1^i \in \mathbb{R}^{\left( \frac{H}{2^i} \cdot \frac{W}{2^i} \right) \times C}$, where C is the number of feature channels. We construct a sparse feature map $a_0^i \in \mathbb{R}^{K \times C}$ according to top-$k$ indices selected from $D_0$. The corresponding area of $A_0^i, A_1^i$ should have higher similarity, and the same intuition also applies to $a_0^i$ and $A_1^i$. Therefore, we can construct a similarity matrix $S_{0\rightarrow 1}\in K\times \left( \frac{H}{2^i} \cdot \frac{W}{2^i} \right)$:
\vspace{-3mm}
\begin{align}
    S_{0\rightarrow 1} = \it Softmax \left(\frac{a_0^i  A_1^i}{\sqrt{C}}\right).
    \label{eq:softmax}
\end{align}
Then we create a coordinate map $B \in \mathbb{R}^{\left(\frac{H}{2^i} \cdot \frac{W}{2^i} \right)\times 2 }$. Using top-$k$ indices selected from $D_0$, we extract the corresponding points from the coordinate map $B$ to form a sparse coordinate map $b_0\in \mathbb{R}^{K \times 2}$. We use the product between similarity matrix $S_{0\rightarrow 1}$ and $b_0$ to represent the estimated position in $I_1$ for selected $k$ points. Therefore, the subtraction between the previous product $S_{0\rightarrow 1}  b_0$ and $b_0$ can represent the flow for those points, namely $f_{0\rightarrow 1}$:
\begin{align}
    \widetilde{f}_{0\rightarrow 1} = S_{0\rightarrow 1}  b_0 - b_0.
\end{align}
Thus, $\widetilde{f}_{0\rightarrow 1}$ is obtained in a global matching manner, making it more capable of capturing large motions. Finally, we reuse the top-$k$ indices from $D_0$ to fill $\widetilde{f}_{0\rightarrow 1}\in \mathbb{R}^{K\times 2}$ to $f_{0\rightarrow 1}\in \mathbb{R}^{\frac{H}{2^i} \times \frac{W}{2^i}\times 2}$ with zero. $f_{1\rightarrow 0}$ can be obtained symmetrically.

\subsection{Flow Shifting}
\label{sec:flow-shifting}
The goal of our sparse global matching branch is to acquire sparse flow compensation $f_{t\rightarrow 0}$ and $f_{t\rightarrow 1}$ to improve the estimated intermediate flows, $\widetilde{F}_{t\rightarrow 0}$, $\widetilde{F}_{t\rightarrow 1}$, with large motion capturing ability. Next, we shift $f_{0\rightarrow 1}$, $f_{1\rightarrow 0}$ to $f_{t\rightarrow 1}$, $f_{t\rightarrow 0}$ by flow shifting, which can help mitigate the coordinate system mismatch.

We intercept a $(1-t)$ proportion of $f_{0\rightarrow 1}$ and shift this $(1-t)$ proportion along the remaining $t$ proportion of the flow $f_{0\rightarrow 1}$. Through this shifting operation, we can obtain $f_{t\rightarrow 1}$, and the shifting operation is performed by forward warping:
\vspace{-2mm}
\begin{align}
    f_{t\rightarrow 1} = \overrightarrow{\mathcal{W}}\left( (1-t) f_{0\rightarrow 1}, t f_{0\rightarrow 1}\right), \label{eq:ft1} \\
    f_{t\rightarrow 0} = \overrightarrow{\mathcal{W}}\left( t f_{1\rightarrow 0}, (1 - t) f_{1\rightarrow 0}\right). \label{eq:ft0}
\end{align}
Note that $f_{t\rightarrow 1}$ is flow shifted by $f_{0\rightarrow 1}$ and $f_{t\rightarrow 0}$ is flow shifted by $f_{1\rightarrow 0}$, we also list the $f_{t\rightarrow 0}$ in \Cref{eq:ft0} for clarity. 

After the forward warping process, the number of sparsely chosen points will potentially change, and we will again choose the top-$k$ points with the least occlusion possibility.

Instead of using the estimated target frame $\widetilde{I}_t$ and input frames $I_0$, $I_1$ to perform global matching and obtain intermediate flow compensation $f_{t\rightarrow 0}$ and $f_{t\rightarrow 1}$, we use global matching between $I_0$, $I_1$ only and perform flow shift. The reason is that $\widetilde{I}_t$ synthesized by $\widetilde{F}_{t\rightarrow 0}$ and $\widetilde{F}_{t\rightarrow 1}$ is not reliable. Since we aim to minimize accumulated propagation errors, we endeavor to minimize the usage of intermediate results wherever possible.

\subsection{Flow Merge Block}
\label{sec:flow-merge}
If we directly replace $\widetilde{F}_{t\rightarrow 0}$ and $\widetilde{F}_{t\rightarrow 1}$ with the flow compensation $f_{t\rightarrow 0}$ and $f_{t\rightarrow 1}$, the resulting flow may lack smoothness. In addition, the flow compensation $f_{t\rightarrow 0}$ and $f_{t\rightarrow 1}$ can also make mistakes. Therefore, we designed a flow merge block $\mathcal{M}$ to adaptively merge $\widetilde{F}_{t\rightarrow 0}$ and $f_{t\rightarrow 0}$, as well as $\widetilde{F}_{t\rightarrow 1}$ and $f_{t\rightarrow 1}$:
 \begin{align}
     \widehat{F}_{t\rightarrow0} = \mathcal{M}(\widetilde{F}_{t\rightarrow 0}, f_{t\rightarrow 0}), \\
    \widehat{F}_{t\rightarrow1} = \mathcal{M}(\widetilde{F}_{t\rightarrow 1}, f_{t\rightarrow 1}).
 \end{align}
 
Inspired by the convex sampling in RAFT~\cite{RAFT}, we take the $\widehat{F}_{t\rightarrow0}$ at each pixel to be the convex combination of $R\times R$ neighbors of $\widetilde{F}_{t\rightarrow 0}$ and $R\times R$ neighbors of $f_{t\rightarrow 0}$, where R is the neighborhood range. We use two convolutional layers to predict a $2\times R^2 \times \frac{H}{2^i}\times \frac{W}{2^i}$ weight assignment, and apply softmax of the $2\times R^2$ neighborhood to obtain masks $\left[W_{main}, W_{sgm}\right]$, where $W_{main}, W_{sgm} \in \mathbb{R}^{R^2 \times \frac{H}{2^i}\times \frac{W}{2^i}}$. $W_{main}$ stands for the weight for the local feature branch flow estimations, and $W_{sgm}$ stands for the weight for sparse global matching block flow compensation. Finally, the flow merge block outputs the $\widehat{F}_{t\rightarrow0}$, obtained by taking weighted sum over the $\widetilde{F}_{t\rightarrow 0}$ and $f_{t\rightarrow 0}$ neighborhood.
\section{Large Motion VFI Benchmark}
\label{sec:benchmark}

To better illustrate our algorithm's effectiveness in handling large motions, we analyzed several widely used large motion datasets, namely, SNU-FILM extreme and hard~\cite{CAIN}, Xiph~\cite{xiph}, and X-Test~\cite{XVFI}. We utilized RAFT~\cite{RAFT} to estimate optical flow between input frames as evidence, allowing for a detailed assessment of motion magnitudes.

According to \Cref{fig:c3}, SNU-FILM and Xiph's mean motion magnitude is much lower than X-Test's. For example, we can see that about $60\%$ of SNU-FILM test cases' mean motion magnitude is below 15, while Xiph has about $40\%$ test cases below 30. 

In addition, we also refer to the criterion in~\cite{LearningTo} for assessing motion sufficiency, requiring at least 10\% of each pixel's flow to have a magnitude of at least 8 pixels within a $270\times 270$ resolution. We adjusted this criterion by lowering the percentage threshold to 5\% while simultaneously increasing the required magnitude to account for the larger resolution of our input frames. This adjusted criterion, representing the minimum magnitude of the top 5\% pixel flows, forms the basis for our proportion ranking of the benchmark datasets in \Cref{fig:c4}.

As depicted in the cumulative distribution chart \Cref{fig:c4}, we found that over 50\% of X-Test have at least 5\% of each pixel's flow with a magnitude of at least 150 pixels. In contrast, Xiph ~\cite{xiph} and SNU-FILM ~\cite{CAIN} contain fewer large motion pixels. Consequently, we focused our evaluation on the most challenging half of these benchmarks to assess our algorithm's performance improvements in handling genuinely large motion data, with details in \Cref{sec:testing}.

\section{Experiments}
\begin{table*}[t]
\centering
\caption{\textbf{Quantitative evaluation (PSNR/SSIM) among different challenging benchmarks} (see \Cref{sec:benchmark}). The best results and the second best results in each column are marked in \textcolor{red}{red} and \textcolor{blue}{blue} respectively. Ours-1/N Points means that we sparsely select 1/N points of the initial intermediate flows estimation to perform global matching by the evidence provided by difference map $D_0, D_1$. ``OOM" denotes the out-of-memory issue when evaluating on an NVIDIA V100-32G GPU.}
\label{tab:main}
\vspace{-3mm}
\begin{tabular}{@{}lcccccc@{}}
\toprule
\multicolumn{1}{c}{\multirow{2}{*}{}} & \multicolumn{2}{c}{X-Test-L}   & \multicolumn{2}{c}{SNU-FILM-L} & \multicolumn{2}{c}{Xiph-L}    \\ \cmidrule(l){2-7} 
\multicolumn{1}{c}{}                  & 2K           & 4K           & hard         & extreme      & 2K           & 4K           \\ \midrule
XVFI~\cite{XVFI}                                  & 29.82/0.8951 & 29.02/0.8866 & 27.58/0.9095 & 22.99/0.8260 & 29.17/0.8449 & 28.09/0.7889 \\
FILM~\cite{FILM}                                  & 30.18/0.8960 & OOM          & 28.38/0.9169 & 23.08/0.8259 & 29.93/0.8537 & 27.14/0.7698 \\
BiFormer~\cite{BiFormer}                              & 30.36/0.9068 & {\color{red}30.14}/{\color{red} 0.9069} & 28.22/0.9154 & 23.57/0.8382 & 29.87/0.8594 & {\color{blue} 29.23}/0.8165 \\
RIFE~\cite{RIFE}                                  & 29.87/0.8805 & 28.98/0.8756 & 28.19/0.9172 & 22.84/0.8230 & 30.18/0.8633 & 28.07/0.7982 \\
EMA-VFI-small~\cite{EMA}                               & 29.51/0.8775 & 28.60/0.8733 & 28.57/0.9189 & 23.18/0.8292 & 30.54/0.8718 & 28.40/0.8109 \\ \midrule
Ours-local-branch                     & 30.39/0.8946 & 29.25/0.8861 & 28.73/0.9207 & 23.19/0.8301 & {\color{blue} 30.89}/0.8745 & 28.59/0.8115 \\
Ours-1/8-Points                       & 30.83/0.9022 & 29.73/0.8928 & 28.82/0.9208 & 23.54/0.8355 & 30.88/0.8749 & 28.90/0.8151 \\
Ours-1/4-Points                       & {\color{blue} 30.88}/{\color{blue} 0.9043} & 29.78/0.8948 & {\color{blue} 28.86}/{\color{blue} 0.9212} & {\color{blue} 23.58}/{\color{blue} 0.8368} & {\color{blue} 30.89}/{\color{blue} 0.8751} & 29.15/{\color{blue} 0.8169} \\
Ours-1/2-Points                       & {\color{red} 30.99}/{\color{red} 0.9072} & {\color{blue} 29.91}/{\color{blue} 0.8972} & {\color{red}28.88}/{\color{red}0.9216} & {\color{red}23.62}/{\color{red}0.8377} & {\color{red}30.93}/{\color{red}0.8755} & {\color{red}29.25}/{\color{red} 0.8180} \\ \bottomrule
\end{tabular}
\end{table*}

\subsection{Implementation Detail}
\textbf{Training Datasets.} We use two training datasets, Vimeo90K~\cite{vimeo} for pretraining on small-to-medium motion and X4K1000FPS (X-Train)~\cite{XVFI} for finetuning on large motion. Vimeo90K contains 51,312 triplets with a resolution of 448x256 for training. It has an average motion magnitude between 1 to 8 pixels~\cite{vimeo}. X4K1000FPS (X-Train) contains 4,408 clips with a resolution of 768x768, each clip has 65 consecutive frames. 

\noindent \textbf{Local Feature Branch Training.} We first train our model framework without the sparse global matching branch on Vimeo90K. We crop each training instance to 256x256 patches and perform random flip, time reversal, and random rotation augmentation, following ~\cite{EMA, RIFE}. The training batch size is set to 32. We use AdamW as our optimizer with $\beta_1=0.9$, $\beta_2=0.999$ and weight decay $1e^{-4}$. After 2,000 warmup steps, we gradually increase the learning rate to $2e^{-4}$, then use cosine annealing for 480k steps (300 epochs) to reduce the learning rate from $2e^{-4}$ to $2e^{-5}$. We follow the training loss design of ~\cite{EMA, RIFE} for both training and finetuning, which is included in \Cref{training}.

\noindent \textbf{Sparse Global Branch Finetuning.} We load our pretrained framework, and a global feature extractor from GMFlow~\cite{gmflow}, operating on $i=3$ in \Cref{eq:softmax}. Then we integrate our sparse global matching branch into the framework, setting the sparsity ratio, i.e. 1/8, 1/4, 1/2, and finetuning the sparse global matching block on X-Train. When fine-tuning, we freeze the pretrained framework and the global feature extractor. We crop $512\times 512$ patches from the original training image, and random resize the input images with 50\% probability to remain the $512\times 512$ size, 25\% probability to downscale to $256\times 256$ size and 25\% probability to $128\times 128$ size. We apply random flipping augmentation, following~\cite{XVFI}. The batch size, learning rate and optimizer are the same as our local feature branch, except that warmup steps are set to 1k steps, and the total steps are set to 13.7k (100 epochs). The parameter freezing and random rescaling are for preserving the model's ability to capture small motion details to the greatest extent possible.

\subsection{Test Dataset}
\label{sec:testing}
The proposed algorithm is designed for capturing large motions. Therefore, we test our model's performance on the most challenging subset of the commonly used large motion benchmark, described in \Cref{sec:benchmark}. We used PSNR (Peak Signal-to-Noise Ratio) and SSIM (Structural Similarity Index) as evaluation metric.

\noindent \textbf{X4K1000FPS (X-Test)}~\cite{XVFI} is a 4K resolution benchmark, with 1000fps high frame rate. X-Test contains 15 clips of 33 successive 4K frames extracted from videos with 1000fps. We have selected X-Test, specifically with the largest temporal gap, naming \textbf{X-Test-L}, as our primary benchmark for evaluating large motion scenarios. We choose the $0_{th}$ and $32_{nd}$ frames as input and evaluate the quality of the synthesized $16_{th}$ output frame. Our testing procedure follows ~\cite{M2M} to provide both 4K and downsampled 2K resolution results.

\noindent \textbf{SNU-FILM}~\cite{CAIN} is a widely-used VFI benchmark, with 1280x720 resolution. It has four different difficulty settings according to the temporal gap between two input frames, each with 310 triplets for evaluation. The larger the temporal gap, the more challenging this benchmark becomes. We test our model on the most challenging half of the SNU-FILM hard and extreme, naming \textbf{SNU-FILM-L}, with 155 triplets each. In the meantime, we also provide the performance on the `easy' and `medium' settings on the whole dataset in \Cref{tab:snueasy} to show our model's capability in handling small-to-medium motions. 

\noindent \textbf{Xiph}~\cite{xiph} is a 4K dataset with 8 scenes, each with 100 consecutive frames, extracted from 60fps videos. While it is often denoted as a benchmark for evaluating large motion, it does not match the level of difficulty exhibited by the X-Test. Therefore, we build this benchmark by doubling the input temporal gap and keeping the most challenging half of the dataset, naming \textbf{Xiph-L}, resulting in 192 test instances. By downsampling and center-cropping, we obtain `Xiph-L-2K' and `Xiph-L-4K' results, following ~\cite{softmax}.

\subsection{Comparison with Previous Methods}
To fully inspect our model's capacity on large motion, we evaluate our model on the aforementioned large motion datasets benchmarks and give our analysis compared results with recent VFI approaches, including ones designed for large motion, such as XVFI~\cite{XVFI}, FILM~\cite{FILM}, BiFormer~\cite{BiFormer}, and ones that performed well on commonly used datasets but not designed for large motion, namely, RIFE~\cite{RIFE}, EMA-VFI~\cite{EMA}. 

As shown in \Cref{tab:main}, the performance of our local feature branch, without finetuned on X-Train, has already surpassed most methods. We attribute this to the capacity of CNN and Transformer hybrid framework and high feature resolution ($H/8 \times W/8$). But without the sparse global matching block, our results are not comparable to BiFormer in SNU-FILM-L extreme, Xiph-L-4K and XVFI in X-Test-L-4K. 

After we incorporate the sparse global matching branch and finetuned on X-Train, our performance on large motion benchmarks are boosted. When we only sparsely select 1/8 of the points in initial flow estimation by the guidance of difference map $D_0, D_1$, our performance improved \textbf{0.44dB} on X-Test-L-2K and \textbf{0.48dB} on X-Test-L-4K in terms of PSNR. With more points introduced to be compensated, the performance can be potentially improved by \textbf{0.6dB} and \textbf{0.66dB} on X-Test-L-2K and X-Test-L-4K respectively. 

\noindent \textbf{Small-to-medium Motion Benchmark Performance.} From the data presented in Table \ref{tab:snueasy}, we observe that our results remain consistent with those in small-to-medium scenarios, and they are comparable to a SOTA algorithm, VFIFormer \cite{VFIFormer}, on these benchmarks. This suggests that our local feature branch already achieves satisfactory results, and introducing our sparse global matching branch has a limited negative impact on the overall performance.
\begin{table}[t]
\centering
\caption{\textbf{Performance on small-to-medium benchmarks, SNU-FILM easy and medium.}}
\label{tab:snueasy}
\begin{tabular}{@{}lcccc@{}}
\toprule
\multirow{3}{*}{} & \multicolumn{4}{c}{SNU-FILM}                           \\ \cmidrule(l){2-5} 
                  & \multicolumn{2}{c}{easy} & \multicolumn{2}{c}{medium} \\ \cmidrule(l){2-5} 
                  & PSNR       & SSIM        & PSNR        & SSIM         \\ \midrule
VFIFormer~\cite{VFIFormer}         & 40.13      & 0.9907      & \textbf{36.09}       & \textbf{0.9799}       \\
Local Branch (ft.)          & \textbf{40.15}      & 0.9907      & 36.07       & 0.9795       \\
1/8 Points        & \textbf{40.15}      & 0.9907      & 36.07       & 0.9795       \\
1/4 Points        & 40.14      & 0.9906      & 36.05       & 0.9795       \\
1/2 Points        & \textbf{40.15}      & 0.9907      & 36.05       & 0.9795       \\ \bottomrule
\end{tabular}
\end{table}

\begin{figure*}[ht]
    \vspace{-6mm}
    \centering
    \includegraphics[width=0.9\linewidth, trim=2cm 7.5cm 1cm 1.5cm, clip=true]{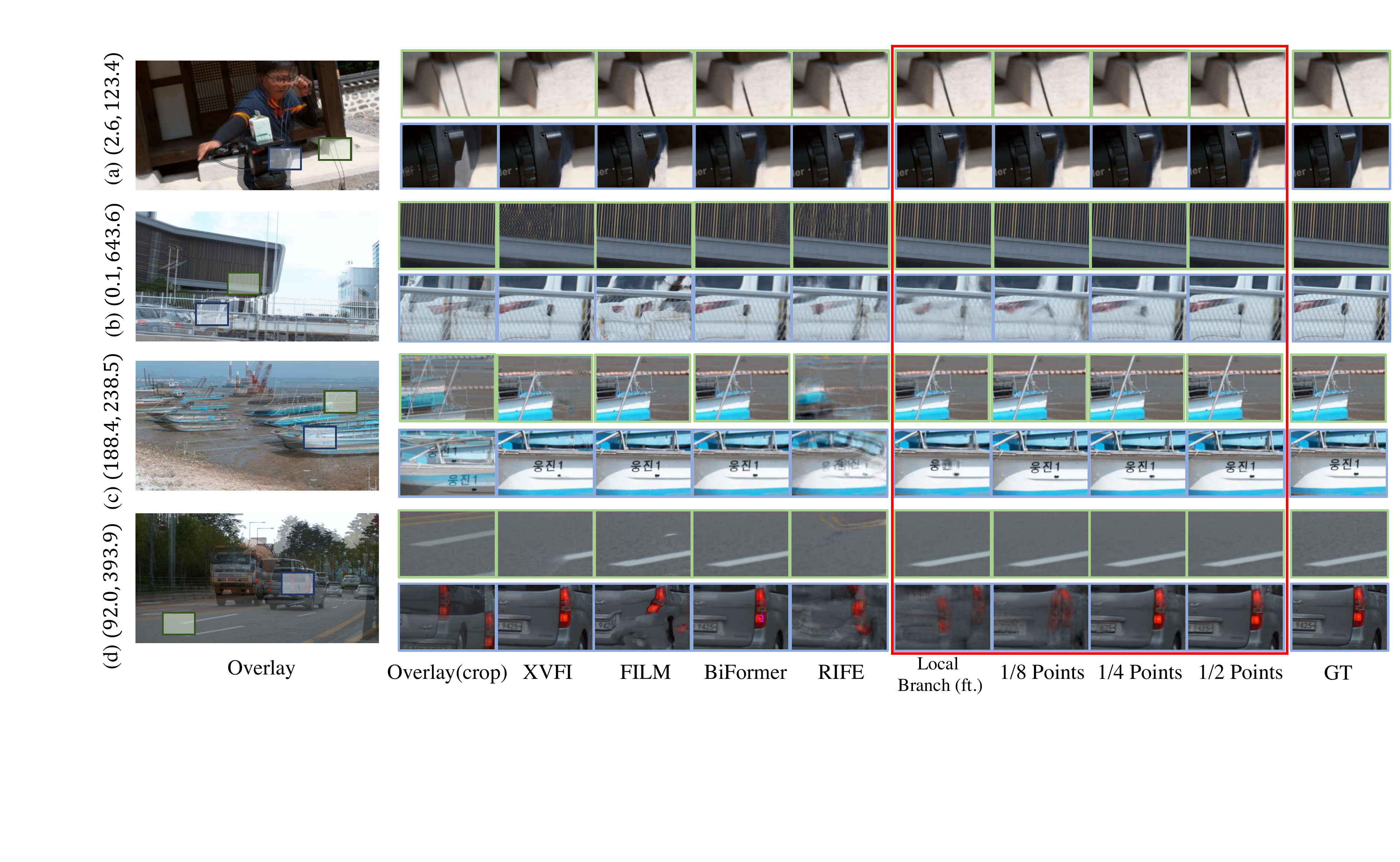}
    \vspace{-3mm}
    \caption{\textbf{Visual comparison with different methods}, instances selected from X-Test-L~\cite{XVFI}. We provide the optical flow magnitude on the left, measured by RAFT~\cite{RAFT}. Four sparsity setting of our methods lies in the \textcolor{red}{red} frame. \textcolor{lightblue}{Blue} frames places a greater emphasis on demonstrating large motion, while \textcolor{lightgreen}{green} frames is more inclined to demonstrate the effect on local details. \textit{Best viewed in zoom.}}
    \label{fig:totalvis}
    \vspace{-3mm}
\end{figure*}

For qualitative results, we also give the visual comparisons between our method and previous VFI methods in \Cref{fig:totalvis}. Four variants of our method lies inside the red frame. For fair comparison, `Local Branch (ft.)' is also finetuned on X-Test, with Flow Refine Block in \Cref{fig:pipeline}, but without sparse global matching modules. In \textcolor{lightblue}{blue} frames, our global compensation branch can fix the unmatched areas with large motion, yielding better visual effect than the `Local Branch (ft.)' model. And with more points added into the sparse global matching block, the visual effect becomes better. 
When compared to other methods, our model can preserve both the \textcolor{lightgreen}{small details} and \textcolor{lightblue}{large motion} well.

\subsection{Ablation Study}
In this section, we ablate the effect of our key component, sparse global matching pipeline fintuning, difference map generation in \Cref{sec:diff-map}, and merge block in \Cref{sec:flow-merge}. 

\noindent \textbf{Sparse Global Matching. } To demonstrate that the primary source of our performance improvements comes from the sparse global matching branch rather than the finetuning process alone, we constructed a Local Branch (ft.) model, incorporating the learnable Flow Refine module (as depicted in Figure \ref{fig:pipeline}), to finetune on the X-Train dataset. Our results indicate that while fine-tuning on a large motion dataset can enhance model performance in challenging scenarios with large motion, global matching consistently outperforms it. Furthermore, we present a comprehensive overview of the global matching performance in Table \ref{tab:baseline}. It is evident that as more points are selected, performance reaches saturation, because not every point needs a global receptive field.
\begin{table}[t]
\vspace{-2mm}
\centering
\caption{\textbf{Comparison of finetuning with different settings.} Local Branch (ft.) contains no global matching.}
\label{tab:baseline}
\begin{tabular}{@{}ccccc@{}}
\toprule
\multirow{2}{*}{}    & \multicolumn{2}{c}{X-Test-L-2K} & \multicolumn{2}{c}{X-Test-L-4K} \\ \cmidrule(l){2-5} 
                     & PSNR         & SSIM          & PSNR         & SSIM          \\ \midrule
Local Branch (ft.)             & 30.58        & 0.8977        & 29.44        & 0.8895        \\
1/8 Points           & 30.83        & 0.9022        & 29.73        & 0.8928        \\
1/4 Points           & 30.88        & 0.9043        & 29.78        & 0.8948        \\
1/2 Points           & 30.99        & 0.9072        & 29.91        & 0.8972        \\
Full Global Matching & \textbf{31.03}        & \textbf{0.9074}        & \textbf{29.95}        & \textbf{0.8974}        \\ \bottomrule
\end{tabular}
\end{table}

\noindent \textbf{Difference Map.} Our sparse global matching algorithm is guided by difference map, $D_0$ and $D_1$, which can help us identify flaws in the initial estimation of intermediate flows. We then select the top-$k$ defective points to correct them. To evaluate the effectiveness of the difference map guidance, we conducted a random sampling strategy to replace the map generation and top-$k$ sampling. The results presented in Table \ref{tab:diff-map} demonstrate that selecting random positions for correction is not as effective as choosing the top-$k$ defective points from the difference map. When dealing with sparse points, random sampling can even decrease accuracy by replacing correct flow with an inaccurate flow compensation. As the number of points increases, this gap narrows. Nevertheless, it is still evident that selecting top-$k$ on the difference map outperforms random sampling.

\begin{table}[t]
\caption{\textbf{Comparison on random sampling $k$ points between generating difference map and sampling top-$k$ points.}}
\vspace{-2mm}
\label{tab:diff-map}
\centering
\begin{tabular}{@{}lllll@{}}
\toprule
\multirow{2}{*}{} & \multicolumn{2}{c}{X-Test-L-2K} & \multicolumn{2}{c}{X-Test-L-4K} \\ \cmidrule(l){2-5} 
                  & PSNR         & SSIM          & PSNR         & SSIM          \\ \midrule
Ours-1/2-top-$k$  & \textbf{30.99}        & \textbf{0.9072}        & \textbf{29.91}        & \textbf{0.8972}        \\
Ours-1/4-top-$k$  & 30.88        & 0.9043        & 29.78        & 0.8948        \\
Ours-1/8-top-$k$  & 30.83        & 0.9022        & 29.73        & 0.8928        \\ \midrule
Local Branch (ft.)          & 30.58        & 0.8977        & 29.44        & 0.8895        \\ \midrule
Ours-1/2-random   & 30.89        & 0.9057        & 29.86        & 0.8967        \\
Ours-1/4-random   & 30.67        & 0.9014        & 29.55        & 0.8923        \\
Ours-1/8-random   & 30.55        & 0.8992        & 29.47        & 0.8893        \\ \bottomrule
\end{tabular}
\end{table}

\noindent \textbf{Flow Merge.} To show the effectiveness of our flow merge block, we remove the merge block and directly apply the sparse flow compensation patches $f_{t\rightarrow 0}$ and $f_{t\rightarrow 1}$ on $\widetilde{F}_{t\rightarrow 0}$ and $\widetilde{F}_{t\rightarrow 1}$, respectively. From \Cref{tab:merge-block}, we can see that after removing the merge block, the obtained results still exhibit slightly higher performance than the fine-tuned local branch model, indicating the effectiveness of our flow compensation patches $f_{t\rightarrow 0}$ and $f_{t\rightarrow 1}$. However, it is noteworthy that as we involve more points in the flow compensation, the improvement in results becomes negligible and even falls short of only using 1/8 points, highlighting the necessity of our merge block.

\begin{table}[t]
\caption{\textbf{Results of our structure with or w/o merge block.}}
\vspace{-2mm}
\label{tab:merge-block}
\centering
\resizebox{\columnwidth}{!}{%
\begin{tabular}{@{}cccccc@{}}
\toprule
\multirow{2}{*}{} & \multirow{2}{*}{Merge Block} & \multicolumn{2}{c}{X-Test-L-2K} & \multicolumn{2}{c}{X-Test-L-4K} \\ \cmidrule(l){3-6} 
         &          & PSNR  & SSIM   & PSNR  & SSIM   \\ \midrule
Local Branch (ft.) & $\times$ & 30.58 & 0.8977 & 29.44 & 0.8895 \\ 
Local Branch (ft.) & $\surd$  & 30.55 & 0.8971 & 29.44 & 0.8889 \\ \midrule
1/8 Points     & $\surd$  & \textbf{30.83} & 0.9022 & \textbf{29.73} & \textbf{0.8995} \\ \midrule
1/2 Points     & $\times$ & 30.75 & \textbf{0.9025} & 29.69 & 0.8936 \\
1/4 Points     & $\times$ & 30.77 & 0.9014 & 29.70 & 0.8924 \\
1/8 Points     & $\times$ & 30.75 & 0.8992 & 29.64 & 0.8910 \\ \bottomrule
\end{tabular}
}
\vspace{-4mm}
\end{table}

\vspace{-3mm}
\section{Conclusion}
In this paper we have presented a sparse global matching algorithm designed to effectively address the challenges posed by large motion in video frame interpolation. We establish a framework that extracts local features for intermediate flow estimation. Then we target the flaws in the initial flow estimation and perform sparse global matching to produce sparse flow compensation across a global receptive field. By adaptively merging initial intermediate flow estimation with sparse global matching flow compensation, we achieve state-of-the-art performance on the most challenging subset of commonly used large motion datasets while keeping the performance on small-to-medium motion benchmark.
\vspace{-2mm}
\vspace{-3mm}
\paragraph {\bf Acknowledgements.} This work is supported by the National Key R$\&$D Program of China (No. 2022ZD0160900), the National Natural Science Foundation of China (No. 62076119, No. 61921006), and Collaborative Innovation Center of Novel Software Technology and Industrialization.

{
    \small
    \bibliographystyle{ieeenat_fullname}
    \bibliography{main}
}

\clearpage
\appendix
\renewcommand\thefigure{\Alph{section}\arabic{figure}}
\renewcommand\thetable{\Alph{section}\arabic{table}}

\section{Local Feature Branch Model Structure}
\label{local structure}
\subsection{Local Feature Extractor}

\begin{figure}[ht]
    \centering
    \includegraphics[width=0.9\linewidth, trim=4cm 5cm 4cm 4cm, clip=true]{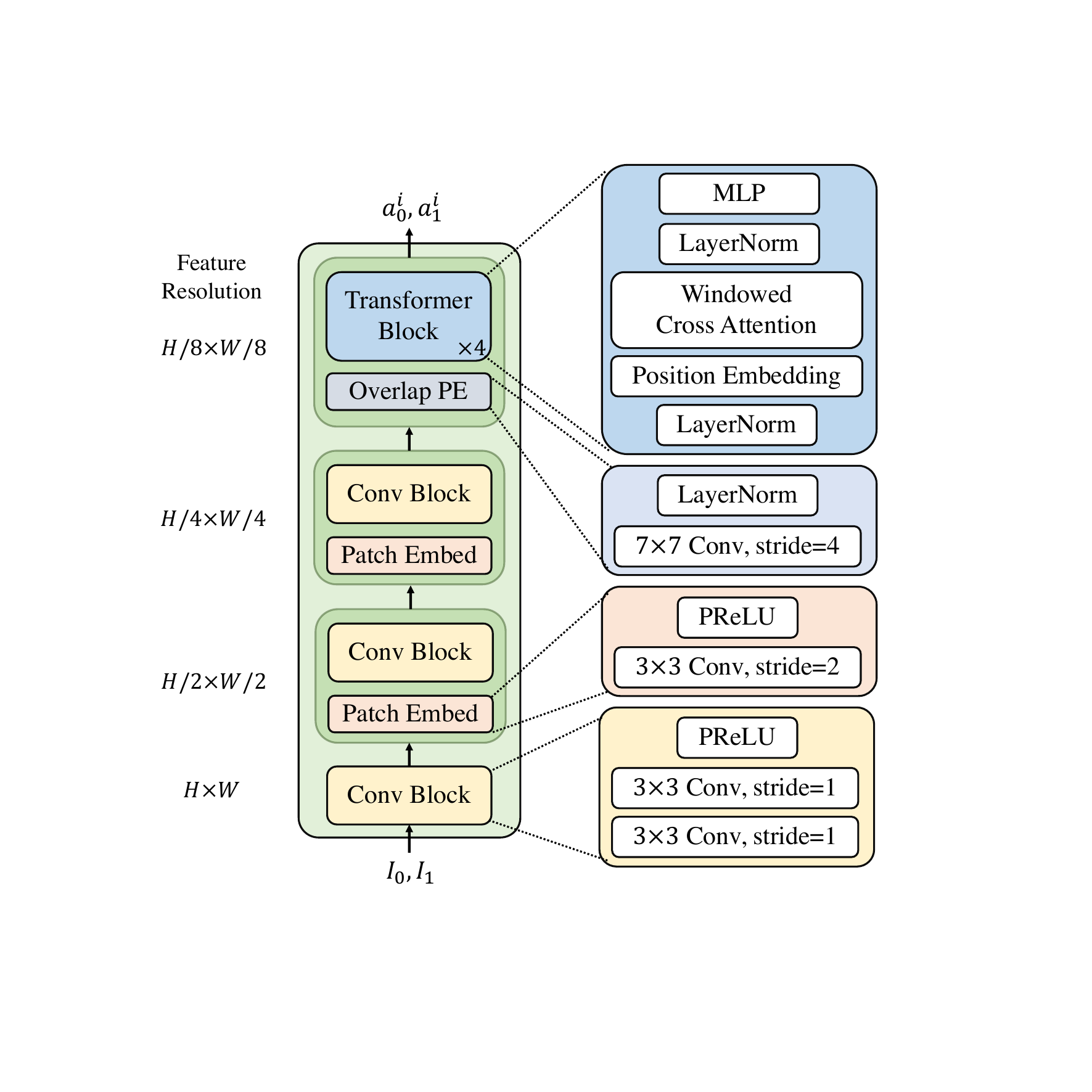}
    \caption{Model Structure of Local Feature Branch. $a^i_0, a^i_1, i\in \{0, 1, 2, 3\}$ is the extracted local feature, corresponding to the feature resolution of $\{H\times W, H/2\times W/2, H/4\times W/4, H/8\times W/8\}$}
    \label{fig:localFeat}
\end{figure}

The structure of our local feature extractor is illustrated in \Cref{fig:localFeat}. As mentioned in \Cref{sec:main-branch}, we adopt a CNN and Transformer hybrid structure for local feature extraction. This design diverges from that of EMA-VFI\cite{EMA} by reducing the network depth. Furthermore, to enhance discriminability within local windows, we incorporate sine-cosine positional embeddings before the windowed cross-attention operation.

\subsection{Flow Estimation}

\begin{figure}[t]
    \centering
    \includegraphics[width=0.6\linewidth, trim=8cm 8cm 8cm 8cm, clip=true]{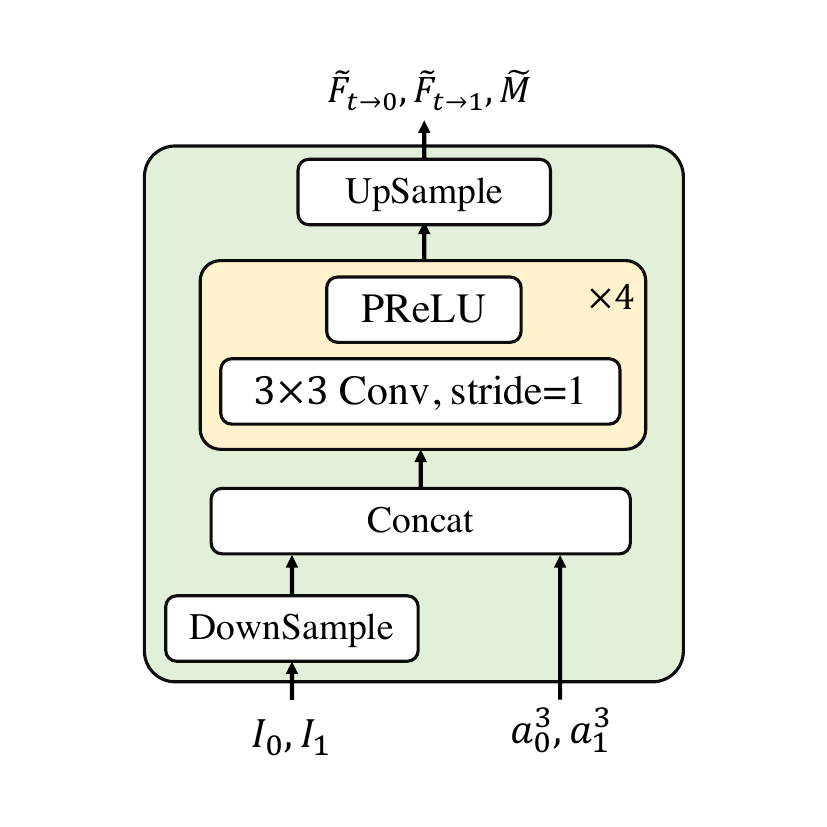}
    \caption{Model Structure of the Initial Flow Estimation Block.}
    \label{fig:motionEst}
\end{figure}

The Flow Estimation Structure, depicted in \Cref{fig:introfig}, consists of two sequential Flow Estimation blocks, as shown in \Cref{fig:motionEst}.  These two blocks are not identical. The first block, detailed in \Cref{fig:motionEst} takes input frames $I_0, I_1 \in H \times W \times 3$ and local features $a_0^3, a_1^3 \in H/8 \times W/8 \times C_3$ as input. Its output includes the initial intermediate flow estimations $\widetilde{F}_{t\rightarrow 0}, \widetilde{F}_{t\rightarrow 1}$, along with the initial fusion map $\widetilde{M}$. 

When pretraining on Vimeo-90K, $\widetilde{F}_{t\rightarrow 0}, \widetilde{F}_{t\rightarrow 1}$ and $\widetilde{M}$ are directly fed into the second block, along with warped images $I_{0\rightarrow t}, I_{1\rightarrow t}$ and the finer local features $a_0^2, a_1^2 \in H/4\times W/4\times C_2$.  In the stages of finetuning and inference, however,  $\widetilde{F}_{t\rightarrow 0}, \widetilde{F}_{t\rightarrow 1}$ and $\widetilde{M}$ are processed by the sparse global matching block for correction, resulting in refined flow estimations $F_{t\rightarrow 0}, F_{t\rightarrow 1}$ and an updated fusion map $M$, which are then input to the second block with $I_{0\rightarrow t}, I_{1\rightarrow t}$ and $a_0^2, a_1^2$.

\subsection{Refine Net}
We follow a similar design in RIFE\cite{RIFE}. We use Context Net to first extract the low-level contextual features. These features are then processed through backward warping, guided by the intermediate flows. The refinement stage involves a U-net shaped network, which can enhance the output frame in a residual form, using the warped features and flows. 
\begin{table*}[ht]
\centering
\caption{\textbf{Results after applying sparse global matching block on EMA-VFI-small.} \textit{1/N} means that we sparsely select 1/N points of the initial intermediate flows estimation. }
\label{tab:ema}
\begin{tabular}{@{}lcccccc@{}}
\toprule
\multirow{2}{*}{} & \multicolumn{2}{c}{X-Test-L}                     & \multicolumn{2}{c}{SNU-FILM-L}                   & \multicolumn{2}{c}{Xiph-L}                      \\ \cmidrule(l){2-7} 
                  & 2K                    & 4K                    & hard                  & extreme               & 2K                    & 4K                    \\ \midrule
EMA-VFI           & 29.51/0.8775          & 28.60/0.8733          & 28.57/0.9189          & 23.18/0.8292          & 30.54/0.8718          & 28.40/0.8109          \\
EMA-VFI-1/8         & 29.65/0.8788          & 28.77/0.8753          & 28.62/0.9192          & 23.31/0.8306          & 30.59/0.8712          & 28.61/0.8114          \\
EMA-VFI-1/4         & 29.81/0.8816          & 28.91/0.8776          & 28.68/0.9196          & 23.41/0.8326          & \textbf{30.64}/0.8720          & 28.78/0.8128          \\
EMA-VFI-1/2         & \textbf{30.12/0.8886}          & \textbf{29.24/0.8840}          & \textbf{28.70/0.9196}          & \textbf{23.46/0.8343}          & 30.63/\textbf{0.8722}          & \textbf{28.91/0.8146}          \\ \bottomrule
\end{tabular}%

\end{table*}

\begin{table*}[ht]
\centering
\caption{\textbf{Results after applying sparse global matching block on RIFE.} \textit{1/N} means that we sparsely select 1/N points of the initial intermediate flows estimation.}
\label{tab:rife}
\begin{tabular}{@{}lcccccc@{}}
\toprule
\multirow{2}{*}{} & \multicolumn{2}{c}{X-Test-L}   & \multicolumn{2}{c}{SNU-FILM-L} & \multicolumn{2}{c}{Xiph-L}    \\ \cmidrule(l){2-7} 
                  & 2K           & 4K           & hard         & extreme      & 2K           & 4K           \\ \midrule
RIFE              & 29.87/0.8805 & 28.98/0.8756 & 28.19/0.9172 & 22.84/0.8230 & 30.18/0.8633 & 28.07/0.7982 \\
RIFE-1/8            & 30.50/0.8902 & 29.52/0.8838 & 28.61/0.9189 & 23.35/0.8298 & 30.26/0.8637 & 28.45/0.8023 \\
RIFE-1/4            & 30.68/0.8981 & 29.72/0.8901 & 28.63/0.9191 & \textbf{23.52}/0.8340 & 30.30/0.8643 & 28.66/0.8048 \\
RIFE-1/2            & \textbf{30.88/0.9034} & \textbf{29.90/0.8944} & \textbf{28.66/0.9195} & \textbf{23.52/0.8350} & \textbf{30.35/0.8656} & \textbf{28.69/0.8066} \\ \bottomrule
\end{tabular}
\end{table*}

\section{Model Loss}
\label{training}
We use the same training loss with EMA-VFI~\cite{EMA}, which is the combination of Laplacian loss and warp loss, defined as:
\begin{align}
    \mathcal{L}=\mathcal{L}_{\text {lap }}+\lambda \sum_i \mathcal{L}_{\text {warp }}^i,
\end{align}
where $\lambda$ is the loss weight for warp loss. Following~\cite{EMA}, we set $\lambda$ = 0.5.

\section{Generalizability}
We apply our sparse global matching block on RIFE\cite{RIFE} and EMA\cite{EMA} to show that our two-step strategy is applicable in more similar flow-based structures. The result is presented in \Cref{tab:rife} and \Cref{tab:ema} accordingly. 

\section{Scalability}
We scaled our model to a bigger model size with 59.3M parameters, basically aligned with EMA-VFI-base~\cite{EMA} which has 65.7M parameters. Results listed in \Cref{tab:basemodel}. From \Cref{tab:basemodel}, we can draw the following conclusion.
 
As more points are incorporated into sparse global matching, the performance gradually saturates. 
This observation is intuitive, considering that not every aspect of the initial estimated flow is inaccurate, nor is every aspect of the global matching flow entirely precise. This is evidenced by \Cref{tab:merge-block}, where the merge block is absent in this ablation. However, upon integrating the merge block (refer to \Cref{tab:baseline}), with more points are involved, up to full global matching, performance still has a little improvement with increased point involvement, meaning that there is still potential for enhancement within the local branch of the smaller model with the help of our merge block.

But when we change our model with a larger local branch with more parameters, the capacity of the local branch becomes stronger. Consequently, it becomes evident that involving all points in global matching leads to performance degradation compared to utilizing only half the points, thus affirming our pursuit of sparsity.

\begin{table}[]
\centering
\caption{\textbf{Results on a larger local branch.} Note that we disable the test-time augmentation when testing for direct comparison.}
\label{tab:basemodel}
\begin{tabular}{@{\quad}ccc@{\quad}}
\toprule
\multirow{2}{*}{}    & \multicolumn{2}{c}{XTest-L-2K}   \\ \cmidrule(l){2-3} 
                     & PSNR           & SSIM            \\ \midrule
EMA-VFI~\cite{EMA}              & 30.85          & 0.9005          \\ \midrule
Ours-local branch    & 30.68          & 0.9010          \\ \midrule
Ours-1/8-Points             & 31.10          & 0.9080          \\
Ours-1/4-Points             & 31.19 & 0.9102 \\
Ours-1/2-Points             & \textbf{31.27}          & \textbf{0.9115}          \\
Full Global Matching & 31.20          & 0.9104          \\ \bottomrule
\end{tabular}
\end{table}

\section{Model Size Comparisons}
We conduct a series of parameters and runtime comparisons on an Nvidia RTX 2080Ti GPU. Illustrated in \Cref{tab:fps}, our local branch is aligned with EMA-VFI-small in terms of runtime and parameters, therefore, we mainly compare our results with EMA-VFI-small model setting.

\begin{table}[]
\centering
\caption{\textbf{Comparisons of model size and corresponding performance.} We only list the X-Test-L-2K results for simplicity.}
\label{tab:fps}
\resizebox{\linewidth}{!}{%
\begin{tabular}{@{}ccccc@{}}
\toprule
\multirow{2}{*}{} & \multirow{2}{*}{\begin{tabular}[c]{@{}c@{}}Inference Time on\\ 512x512 Resolution\end{tabular}} & \multirow{2}{*}{Parameters} & \multicolumn{2}{c}{X-Test-L-2K} \\ \cmidrule(l){4-5} 
                  &                                                                                                 &                             & PSNR         & SSIM          \\ \midrule
RIFE              & \textbf{10ms}                                                                                            & 10M                         & 29.87        & 0.8805        \\
EMA-VFI-small     & 25ms                                                                                            & 14.5M                       & 29.51        & 0.8775        \\
EMA-VFI-base      & 132ms                                                                                           & 65.7M                       & 30.85        & 0.9003        \\
XVFI              & 22ms                                                                                            & \textbf{5.6M}                        & 29.82        & 0.8493        \\
BiFormer          & 59ms                                                                                            & 11M                         & 30.32        & 0.9067        \\ \midrule
Ours-local-branch & 23ms                                                                                            & 15.4M                       & 30.39        & 0.8946        \\
Ours-1/2-Points   & 74ms                                                                                            & 20.8M                       & \textbf{30.99}        & \textbf{0.9075}        \\ \bottomrule
\end{tabular}
}
\end{table}

\section{Different Flow Reversal Teqniques}

\begin{table}[ht]
\centering
\caption{\textbf{Comparisons between different flow reversal techniques.}}
\label{tab:flowreverse}
\resizebox{\linewidth}{!}{%
\begin{tabular}{@{}ccccc@{}}
\toprule
\multicolumn{1}{l}{\multirow{2}{*}{}} & \multicolumn{2}{c}{X-Test-L-2K} & \multicolumn{2}{c}{X-Test-L-4K} \\ \cmidrule(l){2-5} 
\multicolumn{1}{l}{}                  & PSNR         & SSIM          & PSNR         & SSIM          \\ \midrule
flow reversal layer~\cite{QVI}                   & 30.57        & 0.8977        & 29.45        & 0.8886        \\
CFR~\cite{XVFI}                                   & 30.73        & 0.9001        & 29.63        & 0.8913        \\
linear combination~\cite{slomo}                    & 30.69        & 0.9000           & 29.59        & 0.8907        \\
CNN layer                             & 30.18        & 0.8932        & 29.13        & 0.8853        \\
linear reversal                       & 30.70         & 0.9017        & 29.59        & 0.8924        \\
flow shift (Ours-1/8-Points)                 & \textbf{30.83}        & \textbf{0.9022}        & \textbf{29.73}        & \textbf{0.8928}        \\ \bottomrule
\end{tabular}
}
\end{table}

We compare our flow shift strategy with the flow reversal layer in ~\cite{QVI}, complementary flow reversal in ~\cite{XVFI}, linear combination in ~\cite{slomo}, CNN layer and linear reversal on Ours-1/8 setting. Shown by \Cref{tab:flowreverse}, our flow-shifting strategy is the most suitable for sparsely sampled flows.

\section{Interpolating multiple frames into two frames}
We follow the recursive interpolation method in FILM~\cite{FILM} and present our multi-frame interpolation (between two frames) results in \Cref{tab:8x}. 

\begin{table}[]
\centering
\caption{$8\times$ \textbf{Interpolation Results on X-Test (PSNR/SSIM)}.}
\label{tab:8x}
\begin{tabular}{@{}ccc@{}}
\toprule
                           & \multicolumn{2}{c}{X-Test (8$\times$ interpolation)} \\ \cmidrule(l){2-3} 
                           & 2K                        & 4K                       \\ \midrule
EMA-VFI-small-t~\cite{EMA} & 31.75/0.9164              & 30.59/0.9078             \\
RIFE-m~\cite{RIFE}         & 32.23/0.9229              & 31.09/0.9141             \\
FILM~\cite{FILM}           & 31.61/0.9174              & OOM                      \\
Ours-1/2                   & \textbf{32.38/0.9272}     & \textbf{31.35/0.9179}    \\ \bottomrule
\end{tabular}
\end{table}

\section{Finetuning or Training From Scratch}
In our experiments, we conducted training from scratch on the Vimeo-90K~\cite{CAIN} dataset using a sparse global matching block with full global matching. This approach still demonstrated noticeable effects attributed to the global matching process. However, as indicated in \Cref{tab:scratch}, the ability to capture large motion was not on par with the results obtained after finetuning on a dataset with larger motion. Therefore, finetuning on a small batch of large motion datasets (X-Train) is more efficient than training from scratch on a large batch of small motion datasets (Vimeo-90K). This efficiency is evidenced by the reduced number of required training steps, with finetuning necessitating only 13.7k steps as opposed to 480k steps for training from scratch. This finding aligns with the observations reported in FILM~\cite{FILM}, suggesting that large motion datasets can bring large motion capturing ability.

\begin{table}[]
\centering
\caption{\textbf{Comparisons between from scratch and finetuning.} }
\label{tab:scratch}
\begin{tabular}{@{}ccccc@{}}
\toprule
\multirow{2}{*}{}   & \multicolumn{2}{c}{X-Test-L-2K} & \multicolumn{2}{c}{X-Test-L-4K} \\ \cmidrule(l){2-5} 
                    & PSNR         & SSIM          & PSNR         & SSIM          \\ \midrule
Ours-local-branch   & 30.39        & 0.8946        & 29.25        & 0.8861        \\
Global-From Scratch & 30.63        & 0.9012        & 29.61        & 0.8958        \\
Global-Finetuning   & \textbf{31.03}        & \textbf{0.9074}        & \textbf{29.95}        & \textbf{0.8974}        \\ \bottomrule
\end{tabular}
\end{table}

\section{Failed Matching}
\begin{figure}[t]
\centering
\includegraphics[width=\linewidth, trim=2cm 10.5cm 1.5cm 9cm, clip=true]{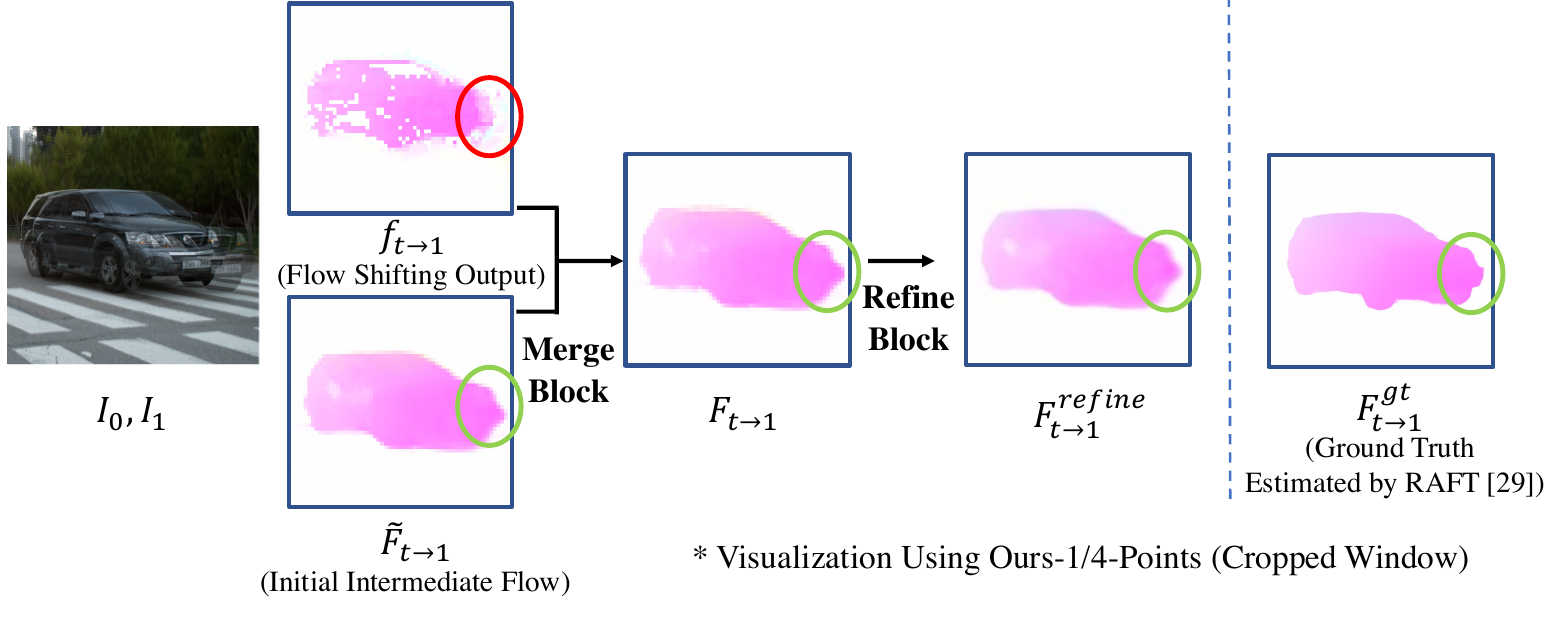}
\caption{Visualization of Matching Failure and Repair}
\label{fig:failed-matching}
\end{figure}

When matching fails, the merge block in our method can adaptively merge the flows, depressing the impact of matching failure. Moreover, we have a refine block to further repair the merged flow. We also provide a visualization in \Cref{fig:failed-matching}.

\section{Inference Speed Bottleneck}
\begin{table}[]
    \centering
    \begin{tabular}{@{}lc@{}}
\toprule
\multicolumn{1}{c}{Operations}                     &  Inference Time   \\ \midrule
Local Feature Branch                               & 23 ms                                                                      \\ \midrule
Flow Compensation Branch                                 & 50.6 ms                                                                    \\
{\color[HTML]{656565} - Global Feature Extraction} & {\color[HTML]{656565} \textbf{45 ms}}                                      \\
{\color[HTML]{656565} - Others}  & {\color[HTML]{656565} 5.6 ms}                                              \\ \midrule
\multicolumn{1}{c}{($512\times 512$ Resolution) Total}                          & 73.6ms                                                                    \\ \bottomrule
\end{tabular}
    \caption{\textbf{Time Profile on Our Proposed Algorithm.} Measured on an Nvidia RTX 2080Ti GPU. }
    \label{tab:op-time}
\end{table}
As shown in \Cref{tab:op-time}, the bottleneck of our pipeline lies in the global feature extractor, instead of other parameter-free components. One naive solution is to replace it with a simpler and lighter global feature extractor in the future. And another solution is to distill the global feature extraction ability from GMFlow~\cite{gmflow} to our own feature extractor, which needs more experiment and probably even training data from optical flow datasets.



\end{document}